# CNN-based Labelled Crack Detection for Image Annotation

Mohsen Asghari Ilani[1,*], Leila Amini[2], Hossein Karimi[3], Maryam Shavali Kuhshuri[4]

**Abstract**

Numerous image processing techniques (IPTs) have been employed to detect crack defects, offering an alternative to human-conducted onsite inspections. These IPTs manipulate images to extract defect features, particularly cracks in surfaces produced through Additive Manufacturing (AM). This article presents a vision-based approach that utilizes deep convolutional neural networks (CNNs) for crack detection in AM surfaces. Traditional image processing techniques face challenges with diverse real-world scenarios and varying crack types. To overcome these challenges, our proposed method leverages CNNs, eliminating the need for extensive feature extraction. Annotation for CNN training is facilitated by LabelImg without the requirement for additional IPTs. The trained CNN, enhanced by OpenCV preprocessing techniques, achieves an outstanding 99.54% accuracy on a dataset of 14,982 annotated images with resolutions of 1536 × 1103 pixels. Evaluation metrics exceeding 96% precision, 98% recall, and a 97% F1-score highlight the precision and effectiveness of the entire process.

## 1. Introduction

Alloys and superalloys often exhibit susceptibility to crack defects, particularly when processed through thermally affected manufacturing methods like laser-based additive manufacturing (AM). These defects encompass solidification cracks, liquation cracks, strain-age cracks, ductility-dip cracks, and cold cracks, as illustrated in ***Figure 1***. The first two are liquid cracks, requiring the presence of liquid films, while the latter three are solid cracks. Due to the intricate thermal history during AM, distinguishing between liquation cracks and solidification cracks, as well as between ductility-dip cracks and strain-age cracks, poses challenges [1]. In our focus on this study, solidification cracks, also known as "hot tears," are emphasized. They are known to occur within the solidifying melt pool or mushy zone when the material is in a semisolid state. Dendrite structures formed during the solidification process impede the flow of remaining liquid in the interdendritic regions, concentrating solidifying stress and leading to the formation of solidification cracks [1, 2]. These cracks exhibit a typical dendritic structure due to the solidification of the liquid, as observed by Carter et al. [3] and Cloots et al. [4]. ***Figure 1*** (**a-f**) displays the morphological

* Mohsen Asghari Ilani
Mohsenasghari1990@ut.ac.ir

[1] School of Mechanical Engineering, College of Engineering, University of Tehran, Tehran, Iran
[2] Department of Information Systems and Business Analytics, College of Business Florida International University, Miami, FL 33199, USA
[3] Department of Computer and Information Technology Engineering, Qazvin Branch, Islamic Azad University, Qazvin, Iran
[4] Department of Electrical Engineering, Faculty of Engineering, University of Isfahan, Isfahan, Iran

characteristics of solidification cracks in PBF-L fabricated Hastelloy X samples, surrounded by grains with different solidification features.

Research by Han et al. [5] indicates that these cracks typically initiate between grains and propagate across pre-solidified layers. Electron backscatter diffraction (EBSD) micrographs reveal crack distribution along paths with high misorientation, particularly high angle grain boundaries (HAGB) exceeding 15°. The solidification behavior of alloys and superalloys with various elements is closely tied to the temperature ranges of their liquidus and solidus temperatures. The critical temperature range (CTR) describes the trend of solidification cracking during the rapid AM solidification process, as depicted in *Figure 1* (**g**). Alloy elements become inhomogeneous within the CTR, resulting in the formation of low-melting films around emerging grains. Marchese et al. [6] observed bright phases along cracks rich in Mo, indicating the formation of Mo-rich carbides and local solute enrichment at grain boundaries. These low-melting liquid films lower the solid–liquid interface energy, promoting cracks through extensive wetting of solid dendrites during the last stage of solidification [7, 8].

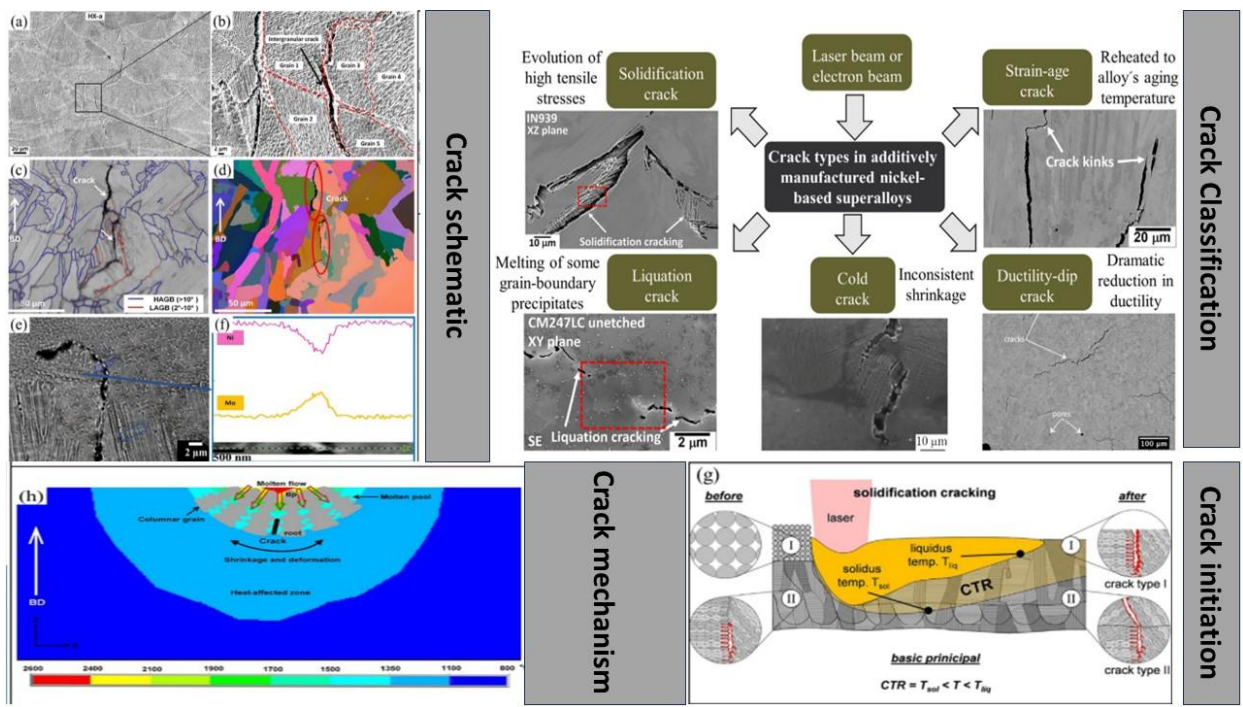

**Figure 1.** Solidification cracks exhibit specific characteristics, as illustrated by various microscopy images. Scanning electron microscopy (SEM) micrographs (**a, b**) and electron backscatter diffraction (EBSD) micrographs (**c, d**) capture cracks in a Hastelloy X sample fabricated through Powder Bed Fusion with L-PBF technology [5]. Additionally, backscattered electron (BSE) imagery and energy-dispersive X-ray spectroscopy (EDS) results of a crack in PBF-fabricated Hastelloy X samples are presented in (**e, f**) [6]. The initiation mechanism of solidification cracks in PBF-fabricated Inconel 738 samples is depicted in (**g**) [4], while the cracking mechanism in a PBF-processed Hastelloy X sample is outlined in (**h**) [9].

As the solidus temperature decreases, brittle grain boundaries may fail to transmit residual tensile stresses or accommodate shrinkage from cooling melt, leading to the separation of adjacent grains and the formation of solidification cracks (crack type II in *Figure 1* (**g**)). Pre-existing solidification cracks can also serve as nuclei for continuous crack growth when partially remelted. Additionally, the liquid pressure drops at the dendrite root, due to insufficient liquid feeding to compensate for solidification shrinkage and deformation, is associated with the formation of solidification cracks (*Figure 1* (**h**)). Bridging between secondary dendrite arms during the critical temperature range impedes liquid feeding, causing solidification cracks to form.

The convolutional neural network (CNN) stands out among existing image processing methods as one of the most capable and promising deep learning approaches for surface classification. Researchers have commonly employed a CNN-based laser crack detection strategy, integrating it into various applications. Notably, the selection of polishing parameters has been informed by predictions of surface conditions made using CNN models. Zhang et al. [10] demonstrated the effectiveness of CNN in predicting and classifying additive manufacturing (AM) surface images. Their work involved classifying surfaces based on the quality of hatch line overlaps to ensure the production of high-quality surfaces. Weimer et al. [11] took a similar approach, developing a deep CNN to automate the detection and classification of defects on machined surfaces. In this case, the CNN systematically extracted relevant features from the training data, allowing for accurate prediction of surface defects with minimal human intervention. The study also compared the performance of various CNN architectures concerning accuracy and computational runtime. Ahmadi et al. [13] developed a supervised machine learning architecture for digital twin applications, focusing on terrain segmentation in coastal regions. By leveraging USGS data and deep learning techniques, this research segments Florida's coastal terrain into distinct classes,a such as water, grassland, and forest, enhancing digital twin models for better environmental monitoring and urban planning.

Building upon the successes of previous CNN-based techniques, it is acknowledged that these methods typically demand a considerable amount of labeled image data for accurate predictions. In response to this challenge, Xiang et al. [12] delved into the development of a semi-supervised CNN model. This innovative approach is tailored to function effectively even when confronted with limited and low-quality labeled data, providing a potential solution to challenges linked to data availability and quality, particularly in certain applications. In the context of this endeavor, the utilization of LabelImg is highlighted as a facilitator in the process. LabelImg plays a crucial role in simplifying the annotation of images, thereby enhancing the understanding of objects in the detection process. Additionally, it contributes to the precise weighting of the region of interest (ROI), enabling a more refined and accurate delineation of the critical elements within the images. This strategic use of LabelImg contributes to overcoming limitations associated with data constraints, ultimately bolstering the performance of the semi-supervised CNN model in scenarios where labeled data is scarce or of lower quality. Ahmadi et al. [24] evaluated SAM and U-Net deep learning models for detecting concrete cracks. SAM effectively identifies longitudinal cracks through image segmentation, while U-Net accurately detects spalling cracks by analyzing pixel characteristics. Using both models together enhances crack detection, which is vital for concrete infrastructure safety and durability. The primary aim of this study is to utilize LabelImg to annotate objects on surfaces featuring numerous defects. The selection of LabelImg as the annotation tool is based on its lightweight design and user-friendly interface, which streamlines the process of labeling object bounding boxes in images. This article provides an overview of LabelImg, highlighting its utility, and offers guidance on annotating images with ease (as shown in *Figure 2*). Recognizing the critical role of image annotation software in computer vision applications, the guide is structured to aid in the evaluation process, assisting in the identification of the most suitable software for both current and future projects. We also reviewed and investigated the application of the segment anything model for defect detection [25], a deeply supervised neural network for classification [26], nanotechnology for material enhancement [27], and behavior detection using trajectory data [28]. Furthermore, we explored stochastic methods for environmental event assessment [29], gradient-based optimization for surface analysis [30], and logistic regression models for drought characterization [31].

The study specifically concentrates on developing a Convolutional Neural Network (CNN) classifier for categorizing additive manufacturing (AM) components based on their surface images [14]. This approach offers a more straightforward and convenient means of automatically identifying surface conditions, providing valuable insights for decisions regarding crack conditions. While existing polishing methods have proven effective in improving the surface quality of metal AM components, they face inherent limitations, primarily related to challenges in controlling the polishing process. To address these challenges, LabelImg is employed in this study to label images based on their geometrical shape, enabling the optimal selection of bounding boxes and annotation labels. This decision is motivated by the complex shapes and varying quantities of defects on the surface, each exhibiting different types. Convolutional Neural networks (CNN) model is used to train and validate the labelled images based on their annotation, bounding boxes and labels.

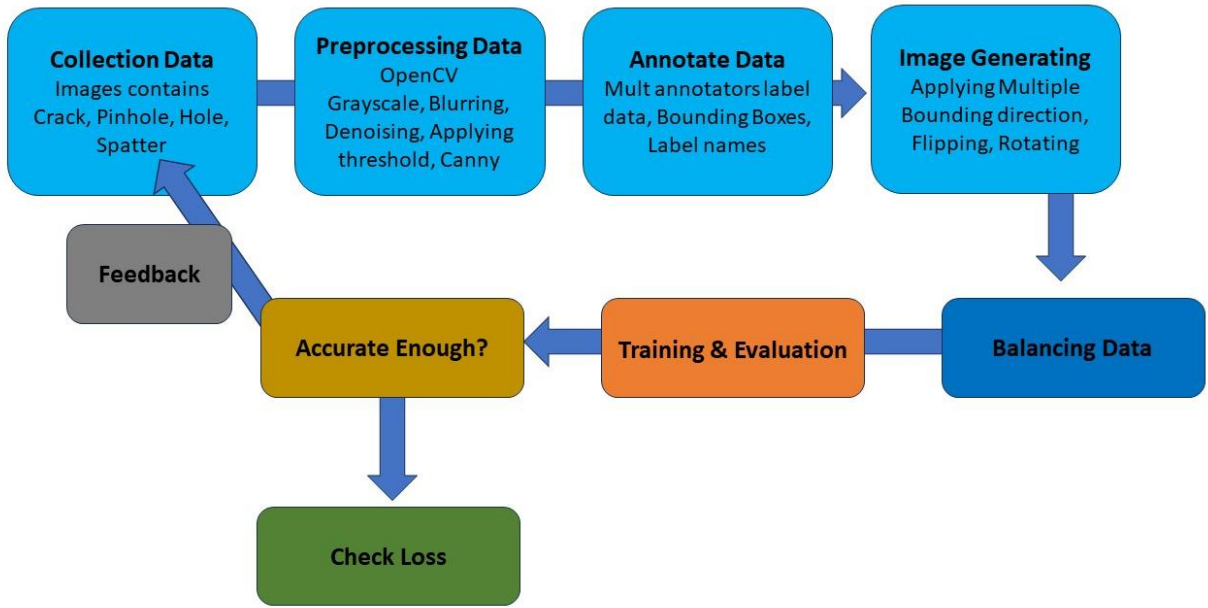

**Figure 2.** LabelImg Approach in Training and Sending Feedback.

## 2. Materials and methods

Convolutional Neural Networks (CNN)

Compared to other machine learning models, a convolutional neural network (CNN) is better equipped for AM surface classification due to its better efficiency, robustness, and generalization capabilities. CNN is a deep learning method which uses inter-connected neurons to classify the images into various groups or categories [15–18]. Each neuron is a computational unit which passes its output to the next layer and so on. Each inter-neuron connection will have a weight and bias associated with it which are tuned during the training phase to maximize the classification accuracy [10, 19]. A significant challenge when working with high-resolution images in CNNs is the demand for higher RAM and the preference for GPU or TPU over CPU. To address this issue, we initially used LabelImg to extract image annotations based on bounding boxes and labels (as depicted in Fig). This step reduces dimensionality for CNN usage. Subsequently, the model was executed on Kaggle utilizing a GPU to accelerate training and leverage a 30GB RAM capacity.

As shown in *Figure 2*, the extracted annotations, post image preprocessing using OpenCV, incorporate features such as bounding boxes and label names in the images, preparing them for training and testing split. Additionally, an image generator function is applied to introduce features in various directions and conditions, enhancing accuracy in the results. In our tailored architecture, the process of feature extraction is directed towards the CNN model, depicted in *Figure 4*. The initial convolution consists of a single layer that takes the input image. Subsequently, four convolutional layers follow with layer sizes of 64, 512, 512, and 256, respectively, concluding with a Maxpool layer. These convolutional layers are succeeded by a maximum pooling layer and a fully connected layer. Each layer utilizes distinct sets of kernels or filters—matrices that traverse the input data to extract relevant features through dot product operations.

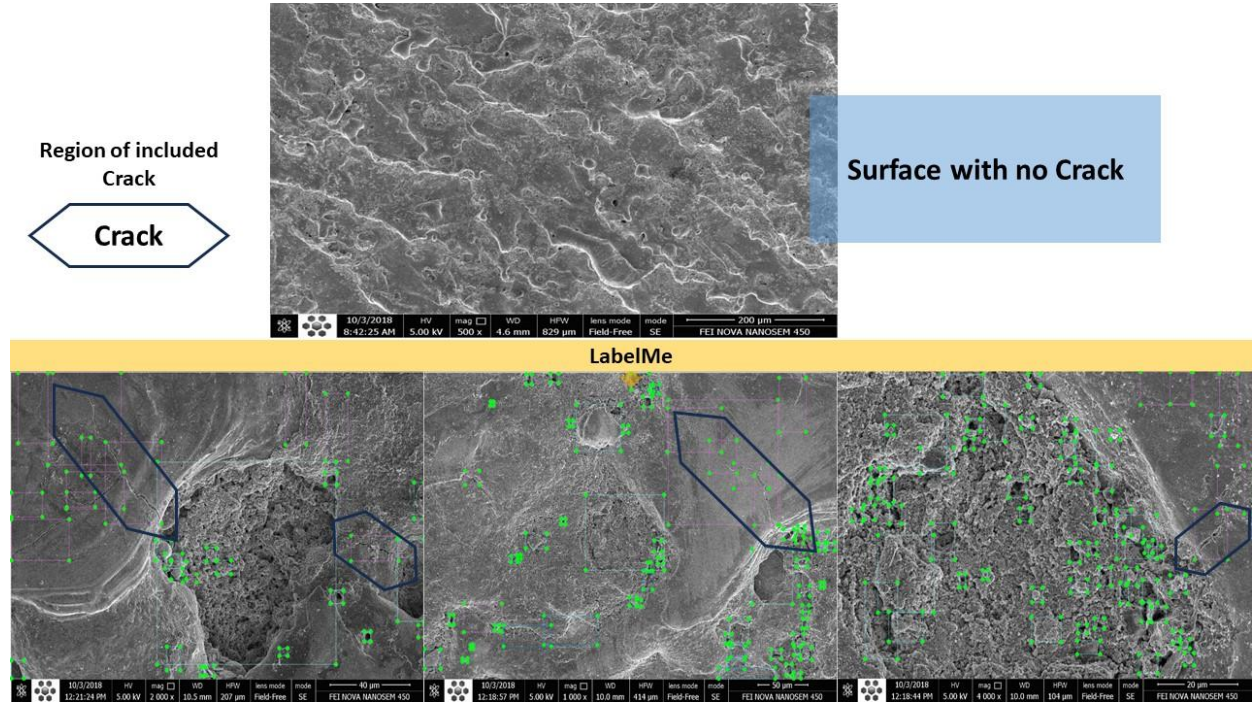

**Figure 3.** Extracting Annotation by LabelImg (Bounding Boxes + Label).

The input image is resized to 80-by-120 pixels. The final activation layer of the CNN structure employs a SoftMax function. This function predicts a multinomial probability distribution based on inputs received from preceding layers. The SoftMax function is expressed as [20]:

$$\sigma(\vec{z})_i = \frac{e^{z_i}}{\sum_{j=1}^{k} e^{z_j}}$$

where $k$ represents the number of classes and $z$ is the input to the final layer.

A cross-entropy loss function is employed for training the neural network. If $\hat{y}_i$ is the predicted class probability, $y_i$ is the true class probability, and $k$ is the number of responses, the cross-entropy loss function is defined as:

$$Loss = -\sum_{i=1}^{k} y_i \cdot \log(\hat{y}_i)$$

The function attains its minimum value when the predicted class probabilities are equal to the true class probabilities.

Cross-entropy stands out as the preferred loss function for image classification tasks, specifically when predicting the probability of a class within a set of classes. Its computational efficiency makes it particularly well-suited for effectively quantifying the difference between actual and predicted probabilities, even in the context of large-scale deep networks. The stability of gradients during backpropagation is a crucial factor in its favor, as incompatible loss functions might generate unstable gradients, leading to training failures. Furthermore, the cross-entropy function is statistically stable, characterized by its convex nature with a single minimum, making it robust against minor input noises. This stability contributes to its effectiveness in training neural networks. The overall summary of the proposed approach is presented in *Figure 4*.

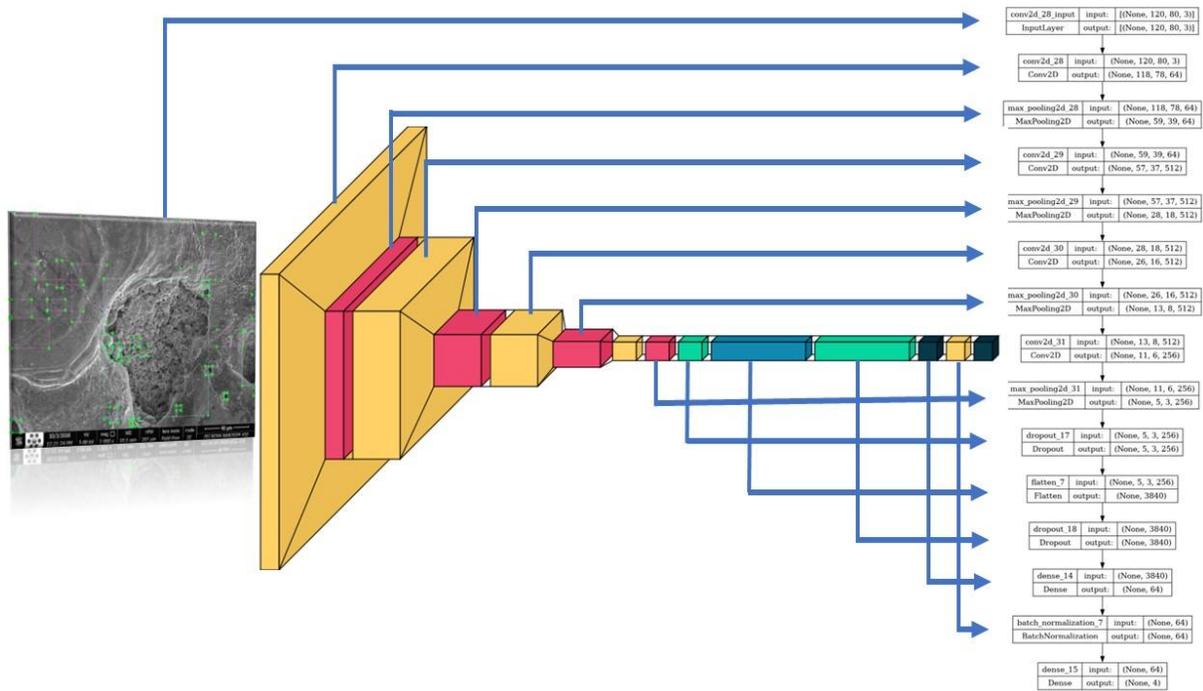

**Figure 4.** Schematic of CNN Architecture.

## 3. Results and Discussion

The assessment of surface crack detection performance in existing literature has primarily relied on private datasets [17, 21–23]. Notably, there is a lack of a publicly available labeled surface crack dataset that includes both bounding box and segmentation parameters. To address this gap, the effectiveness of the proposed architecture is evaluated using a manually collected and labeled dataset specifically tailored for the task of object detection and segmentation. The process involves the use of the "LabelImg" image annotation tool to label cracks within images, storing the parameters of the bounding box and segmentation mask in XML files. These images and associated XML files constitute the training data for the model.

The dataset creation involves the collection and annotation of images to identify surface cracks, incorporating various elements such as lighting effects, shadows, color variations, and background noise. OpenCV libraries are utilized to handle these image variations effectively. To enhance the model's performance, the ImageDataGenerator from TensorFlow is integrated, contributing to a more comprehensive understanding of crack detection. In total, 14,982 annotated images are gathered, subsequently divided into training and test sets to facilitate the detection of surface cracks. This comprehensive dataset aims to contribute to the advancement of surface crack detection methodologies by providing a standardized and openly accessible benchmark for evaluating detection algorithms.

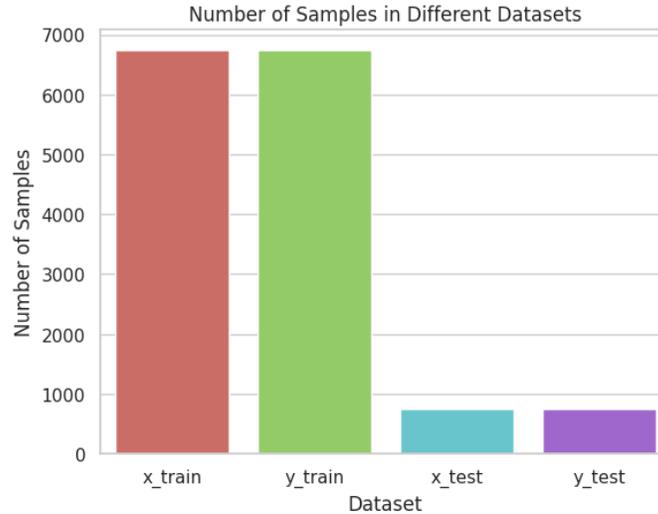

**Figure 5.** Training and Test Datasets.

In contrast to other image-based studies focused on crack detection in additive manufacturing (AM)-built surfaces, this study eschews the need for explicit feature extraction techniques. This decision stems from the inherent capability of Convolutional Neural Networks (CNNs) to autonomously learn features by adjusting the weights of receptive fields during training (LeCun et al., 2012). However, a practical measure is employed for efficient computation, involving the utilization of mean values from the training dataset [23]. Furthermore, the study leverages labeled objects provided by LabelImg, facilitating a more accurate identification of objects compared to relying solely on the model's training. *Figure 4* provides a summary of the training and validation results. The dataset maintains a balanced ratio of crack to intact images at 1:1, with a training-validation split of 4:1. The training accuracy is calculated from a set of 6,742 images, while the validation set comprises 20% of the training data. Remarkably, exceptional accuracy is achieved, reaching the highest values of 99.54% at the 51st epoch for training and 97.95% at the 49th epoch for validation. The use of two GPUs enhances the training speed, reducing the time required to reach the 100th epoch by approximately 90 minutes. However, it's worth noting that the estimated running time on a CPU alone exceeds 6 hours, underscoring the computational efficiency gained through GPU acceleration.

On the contrary, the training loss, as depicted in *Figure 6*, serves as an assessment of the model's adherence to the training data, reflecting its capability to capture patterns within the dataset. In contrast, the validation loss evaluates the model's generalization performance on new, unseen data. The reported values of 0.0033 and 0.0166 for training and validation losses, respectively, signify a heightened accuracy in the fitted model's representation of the underlying patterns in the datasets. These low loss values indicate a robust model that not only effectively learns from the training data but also demonstrates promising performance on previously unseen data, reflecting a capacity for generalization.

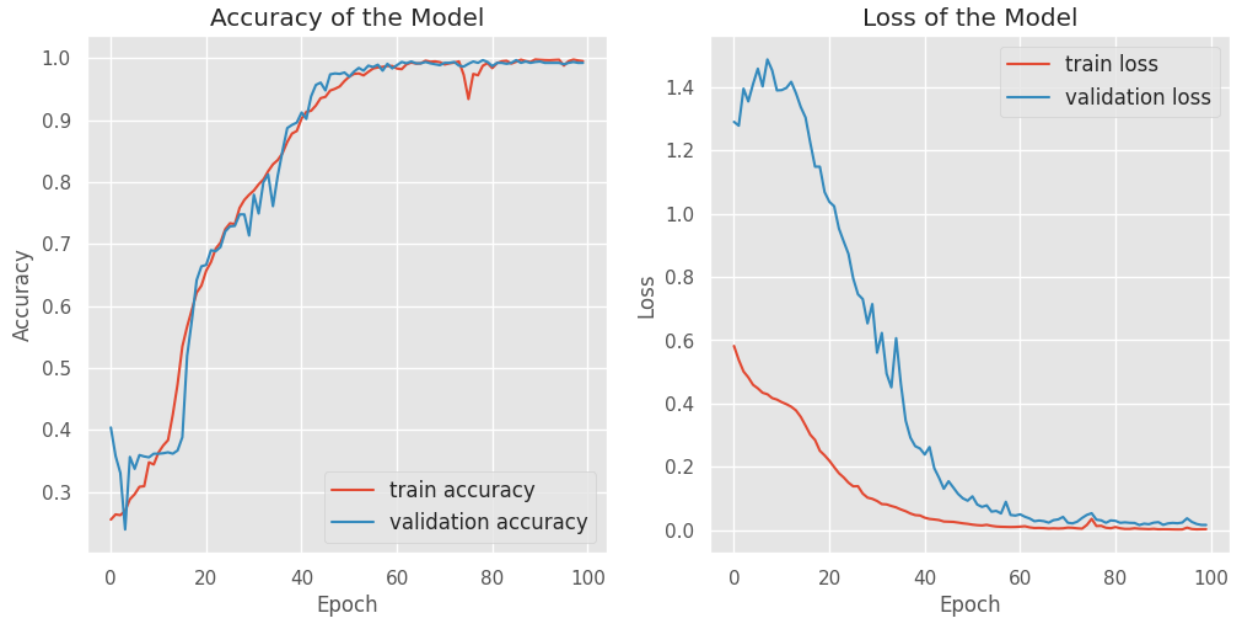

**Figure 6.** Training and Validation Accuracy and Loss per Epoch.

The obtained results from the CNN model demonstrate a highly efficient crack detection performance, as evidenced by the precision, recall, and F1-score metrics for the "Crack" label, as *Table 1*. Specifically:

**Precision**: 0.96 indicates that 96% of instances predicted as "Crack" by the model were indeed true positives, minimizing false positives.

**Recall**: 0.98 signifies that the model captured 98% of the actual "Crack" instances, minimizing false negatives.

**F1-score**: 0.97, a balanced measure of precision and recall, reinforces the effectiveness of the model in correctly identifying "Crack" instances.

These metrics collectively suggest that the CNN model excels in efficiently detecting and accurately classifying of cracks within the dataset. The high precision indicates a low rate of false positives, while the high recall underscores the model's ability to capture a significant portion of actual crack instances. The F1-score reflects a harmonious balance between precision and recall, reinforcing the model's overall efficiency in crack detection.

**Table 1.** Evaluation of CNN-based Model.

| Label | Number | precision | Recall | F1-score | Support |
|---|---|---|---|---|---|
| Crack | 0 | 0.96 | 0.98 | 0.97 | 50 |
| Pinhole | 1 | 0.99 | 1.00 | 1.00 | 124 |
| Hole | 2 | 0.99 | 0.99 | 0.99 | 258 |
| Spatter | 3 | 1.00 | 0.99 | 1.00 | 318 |
| | | | | | |
| Accuracy | | | | 0.99 | 750 |
| Micro Ave | 0.99 | 0.99 | 0.99 | 0.99 | 750 |
| Weighted Ave | 0.99 | 0.99 | 0.99 | 0.99 | 750 |

## 4. Conclusion

The study introduces an innovative LabelImg approach to enhance surface defect detection accuracy in additively manufactured (AM) components using CNN-based intelligence. The Balling phenomena, associated with crack defects, prompts the application of a CNN-based strategy. After defect labeling, the annotation (bounding boxes and labels) is separated for training and testing, also serving as feedback for refining preprocessing hyperparameters. The results emphasize the deep learning approach's effectiveness in precisely categorizing AM-built surfaces, offering significant time and cost savings in industrial applications.

a) Successfully training a CNN model to predict AM component surface categories with over 99% accuracy.
b) The CNN-based model demonstrates complexity in crack types, achieving 96% precision, 98% recall, and a 97% F1-score. Future investigations may explore crack type separation for enhanced precision.
c) Balancing datasets substantially improves training and validation, elevating accuracy from 32% to 99%.
d) Evaluation metrics exceeding 96% precision, 98% recall, and a 97% F1-score highlight the precision and effectiveness of the entire process.
e) LabelImg is recognized as a valuable tool for images with intricate details.

To address the complexity introduced by various crack types, an innovative approach is suggested—recognizing cracks based on types and applying features and physics of the process. This includes leveraging thermal energy emitted by lasers for material melting and vaporization, leading to a proposed physics-informed CNN-based crack detection method for more precise identification. This forward-thinking approach aims to enhance the model's capabilities and broaden its applicability.

**Author contributions** Mohsen Asghari Ilani- Design of manuscript, Writing a manuscript and software;

Leila Amini - Writing a manuscript; Hossein Karimi, Maryam Shavali Kuhshuri- Data analysis.

**Funding** There is no funding details to be mentioned in the manuscript.

**Data availability** There is no need to mention the availability of data and materials in the present study.

**Declarations**

**Ethical approval** There is no ethical approval needed in the present study.

**Consent to participate** There is no consent to participate needed in the present study.

**Consent to Publish** There is no consent to publish needed in the present study.

**Competing Interests** There is no competing interests to be mentioned in the present study.